\pdfoutput=1

\documentclass[11pt]{article}

\usepackage[final]{acl}

\usepackage{times}
\usepackage{latexsym}
\usepackage{bm}
\usepackage{hyperref}
\usepackage{color,colortbl}
\usepackage{xcolor} 
\usepackage{soul}
\usepackage{framed}
\definecolor{shadecolor}{rgb}{0.92,0.92,0.92}
\usepackage{stfloats}
\usepackage{enumitem}
\usepackage{booktabs}
\usepackage{nccmath}
\usepackage{multirow}
\usepackage{latexsym}
\usepackage{todonotes}
\usepackage{makecell}
\usepackage{dashrule}
\usepackage{amssymb}
\usepackage{bbding}
\usepackage{arydshln}
\usepackage{subcaption}

\usepackage[T1]{fontenc}

\usepackage[utf8]{inputenc}

\usepackage{microtype}

\usepackage{inconsolata}

\usepackage{graphicx}

\DeclareRobustCommand{\hlgray}[1]{{\sethlcolor{lightgray!50}\hl{#1}}}

\title{Multidimensional Consistency Improves Reasoning in Language Models}

\author{
Huiyuan Lai$^{1,2}$\thanks{Most work carried out while affiliated with the University of Groningen.}, Xiao Zhang$^3$, Malvina Nissim$^3$\\
$^1$National Key Laboratory for Novel Software Technology, Nanjing University, China \\
$^2$School of Artificial Intelligence, Nanjing University, China \\
$^3$CLCG, University of Groningen, The Netherlands\\
\texttt{h.lai@nju.edu.cn, \{xiao.zhang, m.nissim\}@rug.nl}
}

\begin{document}
\maketitle
\begin{abstract}
While Large language models (LLMs) have proved able to address some complex reasoning tasks, we also know that they are highly sensitive to input variation, which can lead to different solution paths and final answers. Answer consistency across input variations can thus be taken as a sign of stronger confidence. Leveraging this aspect, we introduce  a framework where models are \textit{systematically pushed to diversify solution paths} towards a final answer, thereby testing them for answer consistency across multiple input variations. Focusing on math problems, we induce variations in (i) order of shots in prompt, (ii) problem phrasing, and (iii) languages used. Experiments on a wide range of open-source LLMs of various sizes show that reasoning consistency differs by variation dimension, and that by aggregating consistency across dimensions, our framework enhances mathematical reasoning performance on monolingual datasets GSM8K and MATH500, and the multilingual dataset MGSM. We also discuss the different efficacy of the types of variations induced and how they could be further leveraged in future work. 
\end{abstract}

\section{Introduction}

Large Language Models (LLMs) have shown impressive abilities in addressing a variety of complex reasoning tasks, such as mathematical reasoning~\citep{brown-etal-2020-language} and commonsense reasoning~\citep{bommasani2022opportunities}. The use of Chain-of-Thought (CoT), i.e., breaking down a problem and taking multiple intermediate steps to gradually arrive at the final answer, endows LLMs with even better performances \citep{wei-et-al-2022-chain}.

\begin{figure}
    \centering
\includegraphics[scale=.45]{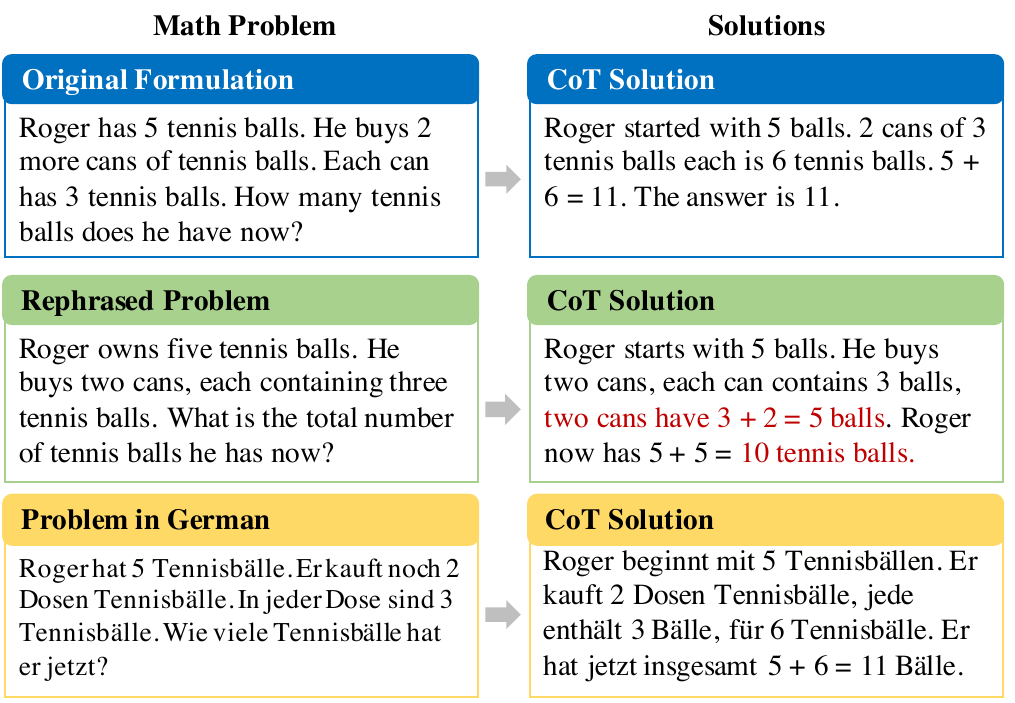}
    \caption{Example of variations: A math problem is presented in different forms or languages, resulting in different reasoning paths and, in some cases, different final answers.}
    \label{fig:overview-idea}
\end{figure}

At the same time, LLMs have also proved to be sensitive and somewhat brittle with respect to variations in the way they are prompted~\citep{zhao-etal-2021-calibrate, lu-etal-2022-fantastically}. For instance, in a few-shot setting for solving mathematical problems, just altering the order in which the example shots are provided might lead to different reasoning paths and possibly different answers~\citep{wang-etal-2022-reationale}; the same can happen if different formulations of the same  problem are used~\citep{zhou-etal-2024-paraphrase}. 
Likewise, an identical mathematical problem presented once in one language, and once in a different one, may be solved following different strategies and also lead to different answers~\citep{lai-nissim-2024-mcot}. Figure~\ref{fig:overview-idea} visualises examples of such variations across different dimensions.

\begin{figure*}[!t]
    \centering
    \includegraphics[scale=.72]{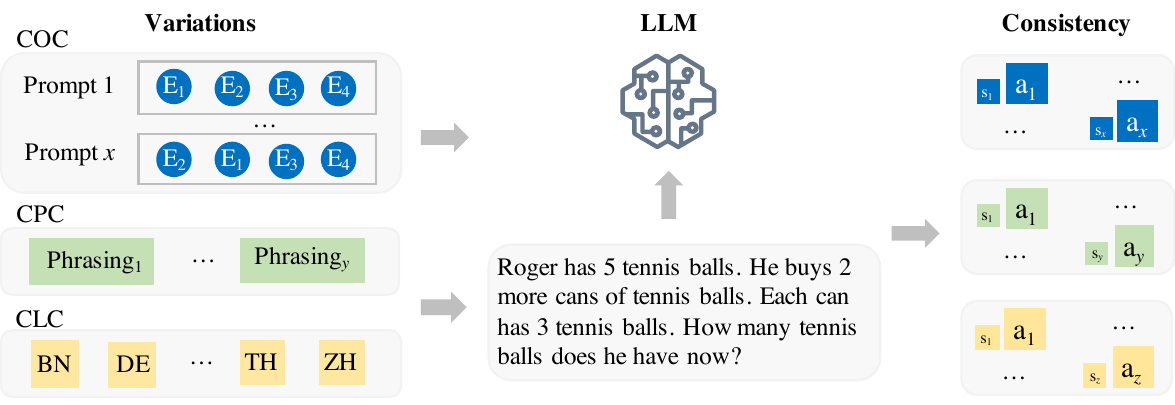}
\caption{Overview of our Multidimensional Reasoning Consistency (MRC) framework: (i) COC changes the exemplars order; (ii) CPC rewrites the given questions in the same language; and (iii) CLC rewrites the given questions in different languages.} 
\label{fig:framework}
\end{figure*}

Some of these variations, such as using even slight alterations in the prompt~\citep{wang-etal-2022-reationale, li-etal-2023-making}, have been exploited in recent work to enhance reasoning performance. 
However, the experimental setup and the assessment of \textit{(in)consistent answers due to variations} are still scattered. In this paper, we argue for a systematic treatment of variations and answer consistency and introduce a {\em Multidimensional Reasoning Consistency} (MRC) framework, focusing on math problems. MRC, shown in Figure~\ref{fig:framework},  allows for a systematic and comprehensive testing and evaluation of model consistency against variations in the way the problem is presented to the model. Our framework also makes it possible to better leverage such variations and answer consistency for improving overall accuracy in mathematical reasoning tasks.

The rationale behind this framework is that by explicitly and systematically pushing the model to likely diversify its solution paths, and possibly yield a different final answer, we can take across-variation consistency of the answer as stronger evidence for its correctness.


We consider three dimensions of variation to test consistency: (i) order of shots in prompt; (ii) problem (re)phrasing; and (iii) natural languages used. We follow~\citet{wang-etal-2022-reationale} in changing the order of the exemplars (i.e., the shots), which results in different prompts based on a set of example problems. 
For problem rephrasing, we prompt the LLMs to rewrite the question before solving it. Lastly, we use the same math problems written in different languages. For each dimension, the LLM generates multiple solution paths to a question, which could differ in various ways, but should in principle lead to the same answer. Answer consistency is eventually used to determine the final answer to the given problem. 

We evaluate MRC on three mathematical reasoning benchmarks: GSM8K~\citep{cobbe-etal-2020-training}, MATH500~\citep{dan2021-measuring}, and MGSM~\citep{shi-etal-2023-language}, covering a range of open-source LLMs with varying scales: 7-8B, 14-32B, and 70-72B. This allows us to systematically examine how model scale affects performance on mathematical reasoning tasks and to compare the effectiveness of our framework across different dimensions of variation.

\paragraph{Contributions} First, we introduce a method to systematically study LLMs' reasoning consistency along multiple dimensions of input variation. Second, we improve model performance on both monolingual and multilingual datasets for a variety of models by leveraging reasoning consistency across variations; this is obtained thanks to the induced substantial diversification of the reasoning paths, offering valuable insights into LLMs reasoning beyond the commonly used sampling-based strategy. 
Third, extensive experimental results show that model consistency differs by variation dimensions, but exploiting consistency always enhances mathematical reasoning performance, and aggregating consistency across dimensions yields an additional boost; 
this paves the way for using a similar framework for other (reasoning) tasks, providing a strategy to make models more robust reasoners. 
All data and code are available at \url{https://github.com/laihuiyuan/MRC}.

\section{Related Work}

\paragraph{Mathematical Reasoning in LLMs}
Mathematical reasoning has garnered great interest in recent times since LLMs have shown what look like complex problem-solving capabilities~\citep{brown-etal-2020-language, lu-etal-2023-survey}. 
With LLMs and few-shot prompting, only a few task examples (e.g., question-answer pair) are required at inference time to enable the LLM to perform the intended task without updating the model parameters~\citep{brown-etal-2020-language}. 
To further elicit LLMs' reasoning capability, \citet{wei-et-al-2022-chain} proposed Chain-of-Thought prompting, which involves an explicit step-by-step reasoning from the question to the answer, rendered in natural language. Given its success, a series of CoT-related methods have been proposed to improve reasoning performance in LLMs, such as complex CoT~\citep{fu-2023-complexity}, auto-CoT~\citep{zhang2023automatic}, 
multilingual CoT~\citep{shi-etal-2023-language}, least-to-most prompting~\citep{zhou2023leasttomost}, 
progressive-hint prompting~\citep{zheng-2023-php}, and residual connection prompting~\citep{jiang2023resprompt}. Rather than developing 
a new specific CoT method, we introduce variations in the prompt and exploit the diversity of CoT outputs. 

\vspace*{-.1cm}
\paragraph{Consistency in LLMs}
In principle, language models could be expected to yield consistent answers in semantically equivalent contexts, especially regarding factual information; this is considered a crucial aspect in assessing model generalization abilities~\citep{fierro-sogaard-2022-factual, lai-nissim-2024-mcot, zhang-etal-2026-knowledgeberg}. In practice, this is often not the case. Some works have thus focused on improving consistency on, e.g., natural language inference~\citep{mitchell-etal-2022-enhancing}, explanation generation~\citep{camburu-etal-2020-make}, cloze test~\citep{ravichander-etal-2020-systematicity}, factual knowledge extraction~\citep{fierro-sogaard-2022-factual}. 
For improving CoT reasoning, \citet{wang2023selfconsistency} suggested to use self-consistency, sampling diverse solution paths and then selecting the most consistent answer.~\citet{zhou-etal-2024-paraphrase} proposed self-consistency-over-paraphrases (SCoP), which diversifies  solution paths by generating different paraphrases for a given problem. To check consistency,~\citet{wang-etal-2022-reationale} use different exemplar orders to possibly trigger diverse solutions.~\citet{lai-nissim-2024-mcot} look at consistency of answers given to the same problem written in two different languages, and use multilingual instruction tuning to improve LLMs' performance across languages. 

Here, we propose a systematic method to study and leverage reasoning consistency along different dimensions to improve overall performance.

\section{Methodology}

Figure~\ref{fig:framework} shows our MRC framework. Using systematic variations, MRC pushes the model to generate multiple solutions for a given question, then measures consistency across variations, and leverages it to improve model performance. 

\subsection{Multidimensional Consistency}

We investigate three variation dimensions that modify different aspects of the model input while preserving the underlying reasoning task, including example order, problem (re)phrasing, and language.

\paragraph{Cross-order Consistency (COC)}
Some prior works have shown that LLMs are sensitive to order, such as the order of options in multiple-choice questions~\citep{pezeshkpour-hruschka-2024-large, zotos-etal-2025-model}, or the order of shots in mathematical reasoning~\citep{wang-etal-2022-reationale}. 
Here we assess how much the \textit{order} of the shots affects consistency of language models. Specifically, we focus on few-shot prompting, which consists of a set of exemplars $($\texttt{question}:$q$, \texttt{step-by-step solution}:$s)$, whose presentation order can be changed arbitrarily. Different permutations of these demonstrations produce multiple semantically equivalent prompts for the same reasoning problem (see Appendix~\ref{sub:prompt} for examples).

\paragraph{Cross-phrasing Consistency (CPC)}
This considers that the surface form of the question itself can also have an impact on the performance of the model.
Differently from~\citet{zhou-etal-2024-paraphrase}, who select 'good' paraphrases based on solving rate improvement as the few-shot exemplars, we directly prompt an LLM to rewrite the question with the goal of making it easier for itself  to solve (see Appendix~\ref{sub:prompt} for examples). We use two main settings: rewrite-without-solve and rewrite-then-solve, which yield the following four different settings when combined with the original question:

\vspace*{-2pt}

\begin{itemize}[leftmargin=*]
\itemsep -0.02in
\item Rewrite-without-solve (RwS): We prompt the LLM to rewrite the question without providing a solution. We then prompt the LLM to solve the rewritten question.

\item Original Question + RwS (RwS+): We concatenate the original question with its rewritten version above and prompt the LLM to generate the solution.

\item Rewrite-then-solve (RtS): We prompt the LLM to first rewrite the question to make it easier to solve and then provide the corresponding solution.

\item RtS Question (RtS-): We prompt the LLM to generate the solution for the rewritten question in the ``rewrite-then-solve'' setting.

\end{itemize}

\paragraph{Cross-lingual Consistency (CLC)} 
One rather outstanding way to vary  formulations is to write the same problem in different languages. Abilities of LLMs in different languages vary substantially, depending on the amount of training data in a given language, and on the similarity of lesser represented languages to more resource-rich ones, as this impacts how well models can deal with less seen languages~\citep{de2022make,muennighoff-etal-2023-crosslingual,ahmet-etal-2024-aya}. 
With cross-lingual consistency, we leverage language diversity to evaluate the LLMs' robustness to input in different languages, and exploit output diversity 
to further improve the LLMs' reasoning performance. Given the same math question in different languages, LLMs are expected to produce reasoning solutions in the corresponding languages. On the one hand, those solutions are expected to arrive at the same final answer if the language model is multilingual; on the other hand, due to the differences in language structures, those solutions can increase diversity compared to using a single language.


\subsection{Reasoning Consistency}
Reasoning stability is reflected by whether a model reaches the same conclusion when solving the same semantically equivalent input. We quantify reasoning stability using Reasoning Consistency (RC),  defined as the agreement among answers generated under different variations of the same input. 

Formally, let $\mathcal{M}$ denote a set of math problems, where each problem is represented as a two-tuple $(\texttt{question}: q, \texttt{answer}: a)$. We define the reasoning consistency of an LLM as the extent to which it produces the same final answer for a given question across a particular dimension of variation (e.g., language). Specifically, for each question, suppose that the LLM generates $n$ candidate solutions $\{s_1, \dots, s_n\}$, which correspond to a set of distinct final answers $\{a_1, \dots, a_m\}$. Reasoning consistency is defined as the proportion of candidate solutions that yield the most frequent final answer:
\begin{align}\label{eq:rc}
    & \textrm{RC(LLM)} = \frac{1} {|\mathcal{M}|}\sum_{i=1}^{|\mathcal{M}|}\frac{\textrm{max}_{j}|\mathcal{S}_{j}|}{n} \\
    & \mathcal{S}_{j} = \{s_i \in \{s_1, \dots, s_{n}\} | f(s_i) = a_j\},
\end{align}

\noindent where $f(s_i)$ maps solution $s_i$ to the final answer. Thus, RC ranges from $1/n$ to $1$, with a higher value indicating greater agreement among the model's answers across variations.

\subsection{MRC for Reasoning}
Eventually, answer consistency across the three dimensions can also be leveraged to improve reasoning performance.  For each question, the language model produces the corresponding solution set $\{s_1, \dots, s_{n}\}$ and the final answer set $\{a_1, \dots, a_{m}\}$. We select the most consistent answer in $n$ solution paths as the final answer 
$\hat{a}$, which is obtained through majority voting:
\begin{equation}
    \label{eq:answer}
    \hat{a} = \arg\max_{a \in \mathcal{A}} \sum_{a' \in \mathcal{A}} \mathbb{I}(a = a')
\end{equation}
\noindent Where $\mathcal{A}$ denotes the set of candidate answers and $\mathbb{I}(\cdot)$ is the indicator function. 

\section{Experimental Setup}

\paragraph{Datasets}
We evaluate our framework on three mathematical reasoning datasets: (i) \textbf{GSM8K}~\citep{cobbe-etal-2020-training}, an English dataset of grade school math word problems (about 7,500 for training and 1,319 for testing); 
(ii) \textbf{MGSM}~\citep{shi-etal-2023-language}, consisting of 250 questions selected from GSM8K and manually translated into ten languages: Bengali (BN), Chinese (ZH), French (FR), German (DE), Japanese (JA), Russian (RU), Spanish (ES), Swahili (SW), Telugu (TE) and Thai (TH). Thus, it contains a total of 11 languages including English; (iii) \textbf{MATH500}, a competition mathematical reasoning benchmark containing 500 problems spanning a range of difficulty levels. Additionally, to test our CLC method on the more challenging dataset, we machine-translate questions from MATH500 into 10 languages in MGSM.\footnote{\url{https://translate.google.com/}.}


\paragraph{Models}
We select a range of open-source state-of-the-art LLMs in varying scales: (i) 7-8B; (ii) 14-32B; and (iii) 70-72B.\footnote{See Appendix~\ref{app:models} for more details on models.} For all models, we only consider instruction-tuned versions.

\paragraph{Implementation}
We perform inference experiments on 4 $\times$ NVIDIA H100 94GB GPUs using the library vLLM~\citep{woosuk-etal-2023-efficient}, without training or fine-tuning language models. During inference, we use few-shot prompts covering the 11 languages released by~\citet{shi-etal-2023-language}. In the multilingual scenario, we use 4-shot for all languages except TE which only uses 2-shot, since a 4-shot prompt would exceed the default maximum length, due to tokenization issues unfavourable to this language~\citep{ahia-etal-2023-languages}. We use greedy decoding unless otherwise specified. For all experiments we report the final answer accuracy except the reasoning consistency score.

\begin{figure*}[t]
    \centering
    \begin{subfigure}[t]{0.49\linewidth}
        \centering
        \includegraphics[width=\linewidth]{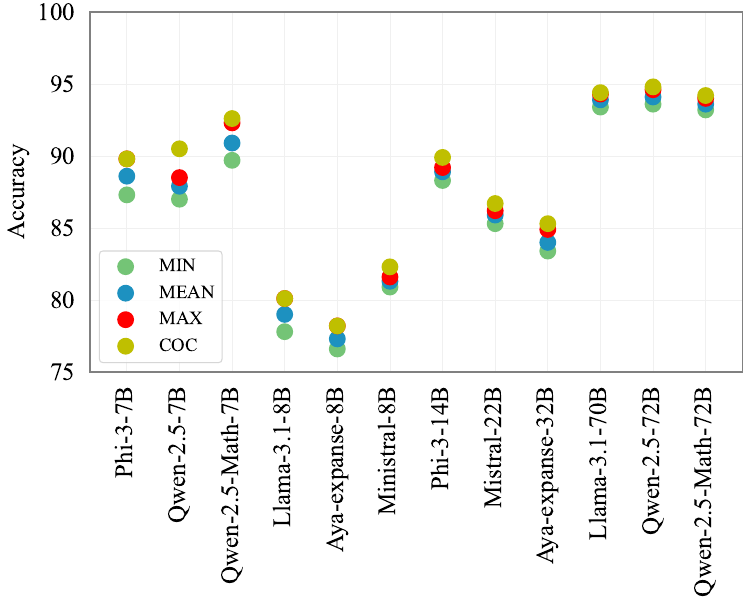}
        \caption{Model performance on GSM8K.}
        \label{fig:coc-gsm8k}
    \end{subfigure}
    \begin{subfigure}[t]{0.49\linewidth}
        \centering
        \includegraphics[width=\linewidth]{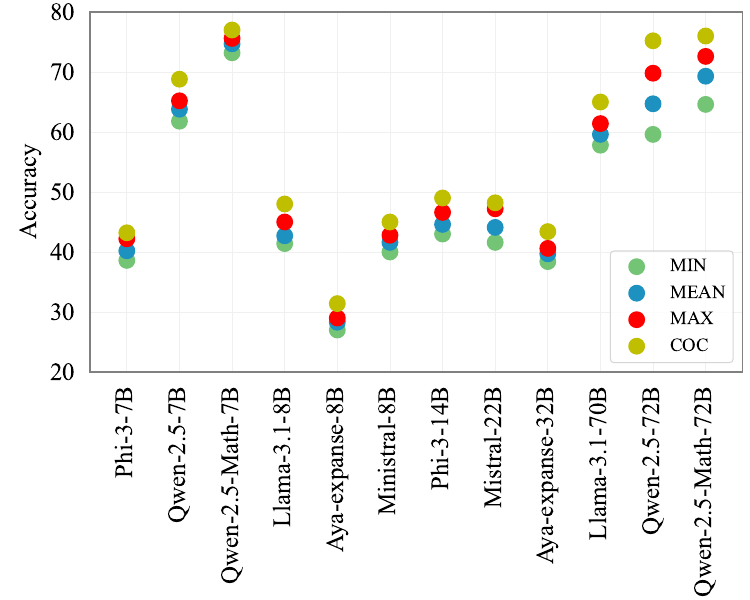}
        \caption{Model performance on MATH500.}
        \label{fig:coc-math500}
    \end{subfigure}
    \caption{Reasoning accuracy of 4-shot for 8 different exemplar orders. Note that MIN, MEAN, and MAX denote the minimum, mean, and maximum scores, respectively.}
    \label{fig:ilc}
\end{figure*}

\begin{figure}[!t]
    \centering
    \includegraphics[scale=.75]{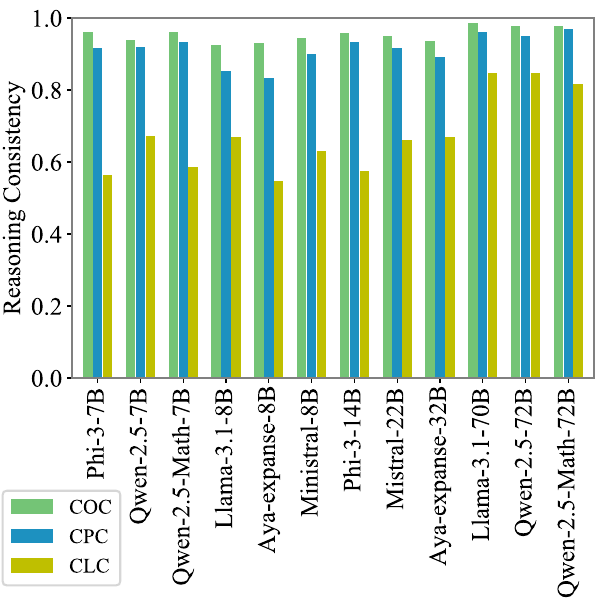}
\caption{Reasoning consistency on three dimensions of variation. Note that COC and CPC are evaluated on the monolingual benchmark GSM8K, while CLC is evaluated on the multilingual benchmark MGSM.} 
\label{fig:consistency}
\end{figure}

\section{Results and Analysis}

We report results for all variation dimensions, and then zoom in on CLC for a more detailed analysis.

\subsection{Reasoning Consistency}
Figure~\ref{fig:consistency} shows reasoning consistency results on the three different dimensions. COC achieves the highest scores, followed by CPC, with CLC having the lowest scores across the board. This suggests that all models are more sensitive to language variations while results are more consistent across different exemplar orders in few-shot prompting. Indeed, when looking at COC only, all models achieve consistency scores above 0.9. Notably, the Llama3.1 family achieves the highest score with the 70B model and the lowest score with the 8B model. 

For CPC and CLC, Aya-expanse-8B has the lowest consistency scores in both dimensions, while larger Qwen2.5 and Llama3.1 models perform best. Compared to COC and CPC, there is a bigger gap in CLC for different models, even within the same scale, e.g., Phi3-7B vs Qwen2.5-7B. Overall, larger models show higher consistency.

\begin{table*}[!t]
\centering
\footnotesize
\setlength{\tabcolsep}{6.5pt}
\resizebox{\linewidth}{!}{%
\begin{tabular}{lcccccc!{\vrule width 0.5pt}cccccc}
\Xhline{1pt}
& \multicolumn{6}{c!{\vrule width 0.5pt}}{\textbf{GSM8K}} & \multicolumn{6}{c}{\textbf{MATH500}}\\
\hline
\textbf{Models} &  \makecell[c]{\textbf{CoT}} & \makecell[c]{\textbf{RwS}} & \makecell[c]{\textbf{RwS+}} & \makecell[c]{\textbf{RtS-}} & \makecell[c]{\textbf{RtS}} & \makecell[c]{\textbf{CPC}} & \makecell[c]{\textbf{CoT}} & \makecell[c]{\textbf{RwS}} & \makecell[c]{\textbf{RwS+}} & \makecell[c]{\textbf{RtS-}} & \makecell[c]{\textbf{RtS}} & \makecell[c]{\textbf{CPC}}\\
\midrule
\hlgray{7-8B}\\
Phi3-7B        & 88.2 & 84.5 & 87.0 & 84.8 & 88.1 & \textbf{90.0} & 40.4 & 36.2 & 41.8 & 39.6 & 43.4 & \textbf{46.6}\\
Qwen2.5-7B      & 88.3 & 86.0 & 89.8 & 86.3 & 90.1 & \textbf{92.0} & 63.6 & 57.2 & 63.6 & 59.0 & 62.0 & \textbf{72.4}\\
Qwen2.5-Math-7B & 90.0 & 87.6 & 91.1 & 89.1 & 92.3 & \textbf{94.1} & 75.2 & 72.8 & 73.2 & 72.4 & 73.2 & \textbf{77.4}\\
Llama3.1-8B    & 79.7 & 73.9 & 78.2 & 77.9 & 81.2 & \textbf{83.8} & 42.6 & 36.6 & 45.6 & 39.0 & 46.6 & \textbf{50.2}\\
Aya-expanse-8B  & 76.7 & 73.5 & 77.9 & 73.5 & 78.2 & \textbf{82.4} & 28.0 & 26.0 & 29.4 & 26.6 & 27.0 & \textbf{32.4}\\
Ministral-8B    & 81.2 & 78.7 & 83.0 & 78.9 & 84.0 & \textbf{84.7} & 42.8 & 41.8 & 48.0 & 39.2 & 50.6 & \textbf{51.0}\\
\midrule
\hlgray{14-32B}\\
Phi3-14B       & 89.2 & 86.4 & 89.2 & 86.9 & 89.8 & \textbf{90.2} & 44.2 & 44.0 & 47.6 & 43.2 & 47.2 &  \textbf{51.2}\\
Mistral-22B     & 85.8 & 83.1 & 85.7 & 84.8 & \textbf{88.1} & \textbf{88.1} & 42.6 & 44.4 & 51.4 & 43.2 & 48.6 &  \textbf{51.6}\\
Aya-expanse-32B & 83.8 & 82.3 & 83.8 & 82.4 & \textbf{88.4} & 88.1 & 38.6 & 37.6 & 38.2 & 41.8 & 40.4 & \textbf{43.2}\\
\midrule
\hlgray{70-72B}\\
Llama3.1-70B   & 94.0 & 89.8 & 93.9 & 91.9 & 93.6 & \textbf{94.8} & 58.6 & 49.0 & 60.2 & 55.8 & 63.4 & \textbf{65.2}\\
Qwen2.5-72B     & 94.6 & 88.9 & 94.4 & 88.6 & 95.5 & \textbf{95.8} & 63.2 & 23.8 & 22.0 & 63.8 & \textbf{74.0} & 67.0\\
Qwen2.5-Math-72B& 94.0& 92.9 & 94.7 & 93.5 & 94.8 & \textbf{95.9} & 66.6 & 54.6 & 58.0 & 67.6 & 74.6 & \textbf{74.8}\\
\Xhline{1pt}
\end{tabular}}
\caption{\label{tab:rewrite-acc}
Reasoning accuracy of CPC on the GSM8K and MATH500 benchmarks, obtained via aggregating vanilla CoT prompting and 4 different question rewriting settings. The best result for each model across the settings is shown in bold.
}
\end{table*}

\begin{table*}[!t]
\centering
\footnotesize
\setlength{\tabcolsep}{7pt}
\resizebox{\linewidth}{!}{%
\begin{tabular}{lrrrrrrrrrrrr}
\toprule
\textbf{Models} &  \makecell[c]{\textbf{BN}} & \makecell[c]{\textbf{DE}} & \makecell[c]{\textbf{EN}} & \makecell[c]{\textbf{ES}} & \makecell[c]{\textbf{FR}} & \makecell[c]{\textbf{JA}} & \makecell[c]{\textbf{RU}} & \makecell[c]{\textbf{SW}} & \makecell[c]{\textbf{TE}} & \makecell[c]{\textbf{TH}} & \makecell[c]{\textbf{ZH}} & \makecell[c]{\textbf{CLC}}\\
\hline
\hlgray{7-8B}\\
Phi3-7B    & 14.8 & 77.6 & 89.2 & 85.2 & 80.4 & 64.8 & 74.4 & 14.0 & 5.2 & 18.8 & 76.0 & \textbf{91.2}\\
Qwen2.5-7B  & 67.2 & 72.4 & 91.6 & 82.8 & 72.0 & 64.8 & 70.8 & 16.4 & 29.2 & 75.6 & 74.0 & \textbf{92.8}\\
Qwen2.5-Math-7B & 16.8 & 76.8 & 92.8 & 82.0 & 76.8 & 61.6 & 78.8 & 4.0 & 5.6 & 51.2 & 85.6 & \textbf{93.6}\\
Llama3.1-8B & 57.6 & 64.4 & \textbf{80.8} & 73.6 & 63.6 & 52.4 & 68.0 & 55.6 & 49.6 & 58.8 & 63.6 & 78.8\\
Aya-expanse-8B & 29.2 & 70.4 & 77.2 & 74.8 & 66.8 & 60.4 & 72.0 & 11.6 & 6.4 & 22.8 & 67.2 & \textbf{82.0}\\
Ministral-8B & 50.4 & 68.0 & \textbf{85.6} & 76.4 & 69.6 & 54.0 & 70.8 & 27.6 & 36.4 & 53.2 & 64.4 & 84.0\\
\hline
\hlgray{14-32B}\\
Phi3-14B   & 14.8 & 76.0 & 88.0 & 87.6 & 76.8 & 72.8 & 80.8 & 18.4 & 5.6 & 12.8 & 77.6 & \textbf{90.0}\\
Mistral-22B & 52.0 & 76.4 & 87.6 & 82.4 & 75.2 & 62.0 & 78.4 & 35.6 & 17.2 & 57.6 & 80.0 & \textbf{89.2}\\
Aya-expanse-32B & 58.4 & 74.0 & 86.0 & 84.4 & 80.0 & 73.6 & 81.2 & 29.2 & 17.2& 52.8 & 77.2 & \textbf{90.8}\\
\hline
\hlgray{70-72B}\\
Llama3.1-70B & 83.6 & 82.0 & 93.6 & 87.6 & 77.6 & 76.8 & 84.4 & 83.2 & 79.2 & 80.4 & 84.0 & \textbf{93.6}\\
Qwen2.5-72B & 88.0 & 84.4 & 93.2 & 88.4 & 80.4 & 84.4 & 87.2 & 66.0 & 68.8 & 91.6 & 86.8 & \textbf{95.6}\\
Qwen2.5-Math-72B & 86.4 & 83.6 & 94.4 & 85.6 & 78.4 & 81.2 & 70.4 & 57.2 & 68.0 & 85.6 & 88.4 & \textbf{95.2}\\
\bottomrule
\end{tabular}}
\caption{\label{tab:clc-gsm8k}
Reasoning accuracy of CLC compared with vanilla CoT prompting on the MGSM benchmark. Bold numbers indicate the best result for each model across individual language settings and the CLC results aggregated across languages.
}
\end{table*}

\subsection{Consistency Improves Reasoning}

For each dimension, we compare the performance obtained exploiting cross-variation consistency to yield a final answer with the performance obtained via the variations on their own.

\paragraph{COC}
Figure~\ref{fig:ilc} reports the results augmented with COC on GSM8K and MATH500, where we use 8 different exemplar orders for the 4-shot prompt.\footnote{Complete results are in Appendix~\ref{tab:coc-results}.} Compared to vanilla CoT prompting, COC improves the reasoning performance for all models. On dataset GSM8K, COC scores are higher than the average scores of 8 different order prompts on all models, and highest on most models, except for Phi3-7B, Llama3.1-8B, and Aya-expanse-8B, where it is on par with the highest scores among the eight ordering configurations we consider in this analysis. On the more challenging dataset MATH500, we see COC achieves the highest scores among all models.

\paragraph{CPC}
Table~\ref{tab:rewrite-acc} presents the results of CPC on GSM8K and MATH500. Accuracy drops when models are fed only the rewritten question (RwS), as they might lose some information from the original question (manual inspection). When combining the rewritten question with the original one (RwS+), most models score comparably to the original prompting and tend to achieve higher scores in the rewrite-then-solve setting (RtS). The latter observation suggests that asking the model to rewrite the question in a simple way and then solve it,  can effectively help the model. Lastly, we see that CPC can further improve the reasoning performance: (i) when comparing to vanilla CoT prompting this is true for all models; and (ii) when comparing to RtS, all models achieve higher accuracy except Aya-expanse-32B on GSM8K and Qwen2.5-72B on MATH500.

\begin{table*}[!t]
\centering
\footnotesize
\setlength{\tabcolsep}{7pt}
\resizebox{\linewidth}{!}{%
\begin{tabular}{lrrrrrrrrrrrr}
\toprule
\textbf{Models} &  \makecell[c]{\textbf{BN}} & \makecell[c]{\textbf{DE}} & \makecell[c]{\textbf{EN}} & \makecell[c]{\textbf{ES}} & \makecell[c]{\textbf{FR}} & \makecell[c]{\textbf{JA}} & \makecell[c]{\textbf{RU}} & \makecell[c]{\textbf{SW}} & \makecell[c]{\textbf{TE}} & \makecell[c]{\textbf{TH}} & \makecell[c]{\textbf{ZH}} & \makecell[c]{\textbf{CLC}}\\
\hline
\hlgray{7-8B}\\
Phi3-7B         & 10.8 & 34.8 & 40.4 & 39.8 & 5.4 & 20.8 & 30.0 & 9.4 & 5.6 & 10.0 & 33.2 & \textbf{44.0}\\
Qwen2.5-7B      & 33.4 & 51.6 & 63.6 & 50.0 & 33.0 & 50.2 & 55.8 & 17.6 & 17.0 & 42.2 & 49.0 & \textbf{67.4}\\
Qwen2.5-Math-7B & 29.2 & 61.2 & 75.2 & 62.0 & 53.4 & 50.8 & 65.8 & 3.8 & 16.2 & 23.6 & 70.6 & \textbf{79.4}\\
Llama3.1-8B     & 22.2 & 31.0 & 42.6 & 35.4 & 7.2 & 29.4 & 38.6 & 24.8 & 9.4 & 26.8 & 32.8 & \textbf{47.2}\\
Aya-expanse-8B   & 10.6 & 25.8 & 28.0 & 26.6 & 25.6 & 24.0 & 25.2 & 8.6 & 5.6 & 11.0 & 24.2 & \textbf{31.2}\\
Ministral-8B     & 17.0 & 31.0 & 42.8 & 42.4 & 36.8 & 26.6 & 38.8 & 8.4 & 7.6 & 18.2 & 29.8 & \textbf{44.4}\\
\hline
\hlgray{14-32B}\\
Phi3-14B        & 3.0 & 42.2 & 44.2 & 41.8 & 9.2 & 33.6 & 37.2 & 13.8 & 4.6 & 1.8 & 38.6 & \textbf{49.6}\\
Mistral-22B      & 16.0 & 43.2 & 42.6 & 43.0 & 23.6 & 34.4 & 43.2 & 15.2 & 35.8 & 23.6 & 35.2 & \textbf{53.8}\\
Aya-expanse-32B  & 20.8 & 40.2 & 38.6 & 38.2 & 3.0 & 35.0 & 38.8 & 15.0 & 10.2 & 16.0 & 38.0 & \textbf{45.0}\\
\hline
\hlgray{70-72B}\\
Llama3.1-70B    & 32.6 & 41.2 & 58.6 & 51.8 & 9.6 & 33.4 & 47.2 & 44.0 & 22.4 & 45.4 & 42.8 & \textbf{64.4}\\
Qwen2.5-72B     & 51.4 & 54.8 & 63.2 & 48.4 & 51.6 & 27.6 & 38.4 & 35.6 & 37.4 & 56.2 & 63.2 & \textbf{75.2}\\
Qwen2.5-Math-72B & 58.0 & 50.0 & 66.6 & 60.0 & 22.4 & 56.6 & 60.6 & 30.6 & 34.4 & 58.0 & 64.4 & \textbf{75.4}\\
\bottomrule
\end{tabular}}
\caption{\label{tab:clc-math500}
Reasoning accuracy of CLC compared to vanilla CoT prompting on the machine-translated MATH500. Note that bold numbers indicate the best result for each model among different languages and CLC.
}
\end{table*}

\paragraph{CLC}
Table~\ref{tab:clc-gsm8k} and Table~\ref{tab:clc-math500} present the results of CLC compared to vanilla CoT prompting on MGSM and machine-translated MATH500, respectively. All models perform best on English, with a serious performance gap between underrepresented (e.g., SW) and high-resource languages, especially for smaller models.
Similar to COC and CPC, compared to vanilla CoT, CLC yields improvement for most models, with Aya-expanse-32B on MGSM, for example, showing a substantial gain of 4.8 percentage points over English. For Llama3.1-8B and Ministral-8B, the accuracy of CLC is slightly lower than that of English, but better than that of all other languages. On the more challenging dataset MATH500, CLC leads to consistent improvements across all models, with the most substantial gain observed on the larger model (e.g, a 12.0 percentage-point absolute improvement over English on Qwen2.5-72B), highlighting its effectiveness in handling more complex tasks and scaling with model size.

\begin{table*}[!t]
\centering
\footnotesize
\setlength{\tabcolsep}{4pt}
\resizebox{\linewidth}{!}{%
\begin{tabular}{lccccccc!{\vrule width 0.5pt}ccccccc}
\Xhline{1pt}
\textbf{Datasets} & \multicolumn{7}{c!{\vrule width 0.5pt}}{\textbf{MGSM}} & \multicolumn{7}{c}{\textbf{MATH500}}\\
\hline
\textbf{Models} &  \makecell[c]{\textbf{COC}} & \makecell[c]{\textbf{CPC}} & \makecell[c]{\textbf{CLC}} & \makecell[c]{\textbf{O\&P}} & \makecell[c]{\textbf{O\&L}} & \makecell[c]{\textbf{P\&L}} & \makecell[c]{\textbf{MRC}} &  \makecell[c]{\textbf{COC}} & \makecell[c]{\textbf{CPC}} & \makecell[c]{\textbf{CLC}} & \makecell[c]{\textbf{O\&P}} & \makecell[c]{\textbf{O\&L}} & \makecell[c]{\textbf{P\&L}} & \makecell[c]{\textbf{MRC}}\\
\hline
\hlgray{7-8B}\\
Phi3-7B        & 92.4 & 92.0 & 91.2 & 93.2 & 93.2 & 93.2 & \textbf{94.8} & 43.2 & 46.6 & 44.0 & 49.0 & 47.6 & \textbf{51.2} & 49.6\\
Qwen2.5-7B     & 92.0 & 93.2 & 92.8 & 93.2 & \textbf{93.6} & \textbf{93.6} & \textbf{93.6} & 68.8 & 72.4 & 67.4 & 72.6 & 72.4 & 73.4 & \textbf{74.0}\\
Qwen2.5-Math-7B & 94.4 & 94.8 & 93.6 & 94.4 & 95.6 & \textbf{96.0} & \textbf{96.0} & 77.0 & 77.4 & 79.4 & 78.6 & 79.6 & \textbf{79.8} & \textbf{79.8}\\
Llama3.1-8B    & 80.8 & 85.6 & 78.8 & 83.6 & 84.0 & \textbf{87.6} & 84.8 & 48.0 & 50.2 & 47.2 & 51.6 & 51.4 & 53.0 & \textbf{53.6}\\
Aya-expanse-8B  & 78.4 & \textbf{85.2} & 82.0 & 80.8 & 79.6 & 84.4 & 82.8 & 31.4 & 32.4 & 31.2 & 33.4 & 33.4 & \textbf{35.2} & 34.2\\
Ministral-8B    & 84.4 & 86.8 & 84.0 & 86.0 & 86.0 & 87.2 & \textbf{87.6} & 45.0 & 51.0 & 44.4 & 47.8 & 47.2 & \textbf{53.2} & 49.6\\
\hline
\hlgray{14-32B}\\
Phi3-14B       & 92.0 & 92.0 & 90.0 & 92.0 & 92.0 & 92.4 & \textbf{93.2} & 49.0 & 51.2 & 49.6 & 51.0 & 52.4 & \textbf{55.6} & 53.8\\
Mistral-22B     & 87.2 & 89.6 & 89.2 & 87.6 & 90.0 & \textbf{93.2} & 91.2 & 48.2 & 51.6 & 53.8 & 52.0 & 53.0 & \textbf{57.4} & 55.4\\
Aya-expanse-32B & 86.0 & 89.2 & 90.8 & 86.4 & 89.6 & \textbf{92.8} & 90.0 & 43.4 & 43.2 & 45.0 & 45.6 & 46.0 & \textbf{47.8} & 47.4\\
\hline
\hlgray{70-72B}\\
Llama3.1-70B    & 95.6 & \textbf{96.8} & 93.6 & 96.0 & 96.4 & \textbf{96.8} & 96.4 & 65.0 & 65.2 & 64.4 & 68.8 & 68.2 & 68.0 & \textbf{69.0}\\
Qwen2.5-72B     & 96.0 & \textbf{97.6} & 95.6 & 97.2 & 96.8 & 97.2 & 96.8 & 75.2 & 67.0 & 75.2 & 77.6 & 77.6 & 76.6 & \textbf{78.8}\\
Qwen2.5-Math-72B& 94.4 & \textbf{95.2} & \textbf{95.2} & 94.8 & 94.8 & 94.8 & \textbf{95.2} & 76.0 & 74.8 & 75.4 & 78.6 & 79.4 & 79.0 & \textbf{80.6}\\
\Xhline{1pt}
\end{tabular}}
\caption{\label{tab:mrc}
Accuracy on MGSM and MATH500. Notes: (i) CPC uses 5 solution paths, COC and CLC use 8 each, and CLC uses 8 languages (excl. BN, SW, TE), so MRC contains a total of 19 paths (excluding the two identical English paths); (ii) best result for each model is bolded.
}
\end{table*}

\paragraph{Cross-dimension Consistency}
Table~\ref{tab:mrc} shows ablation results of different combinations on MGSM and machine-translated MATH500. Of the three variation dimensions, CPC performs best overall, followed by COC and CLC. This suggests that CPC can push the model to better diversify its solution paths, while for CLC, this might be due to the large performance gap between English and other languages. 
The combination of CPC and CLC (P\&L) consistently provides strong performance, even outperforming MRC (aggregating three dimensions) in some models, highlighting their novel and impactful role in our framework. While integrating COC occasionally fails to yield additional benefits, bear in mind that each dimension—when applied individually—still brings improvements. Therefore, promoting reasoning consistency across variations can still help induce substantial diversification of the reasoning paths, offering valuable insights into LLM reasoning beyond commonly used sampling-based strategies. 


\begin{figure*}[!t]
    \centering
    \includegraphics[scale=.58]{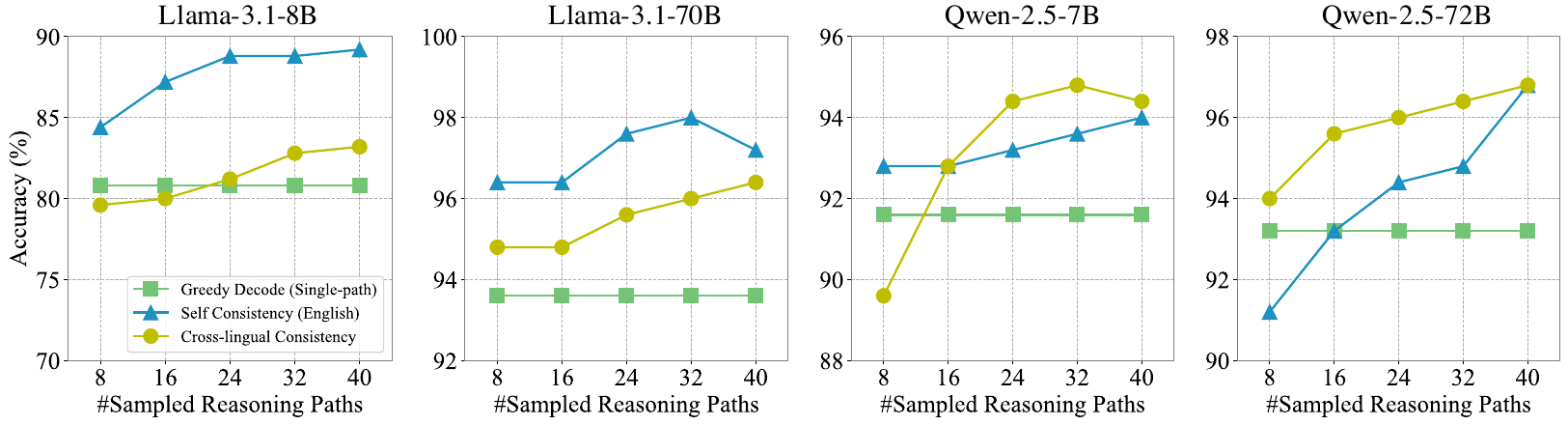}
\caption{Reasoning accuracy of using varying numbers of reasoning paths on GSM8K.} 
\label{fig:varying-paths}
\end{figure*}

\begin{figure*}[!t]
    \centering
    \includegraphics[scale=.72]{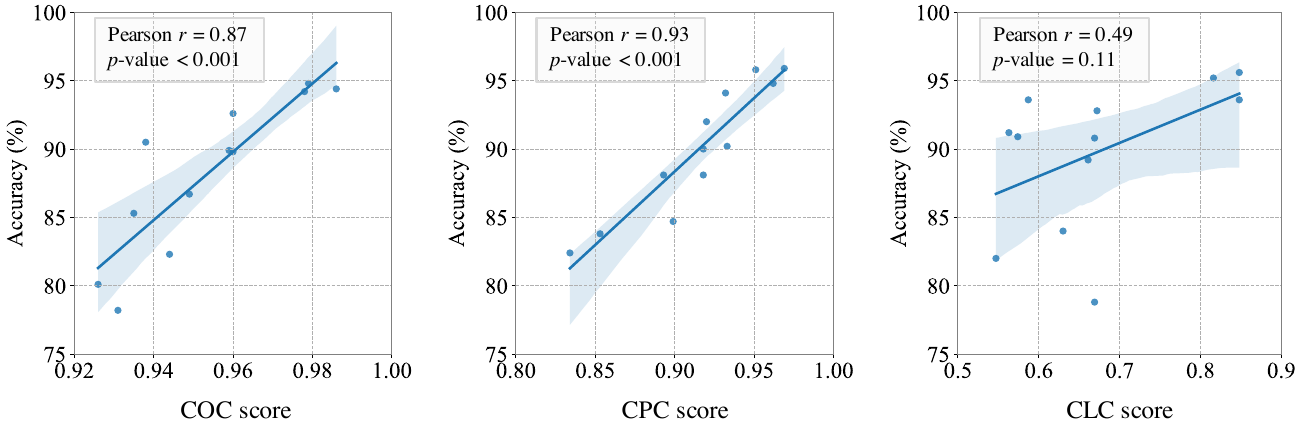}
\caption{Pearson correlation between models' accuracy and different consistency scores.} 
\label{fig:cor}
\end{figure*}

\begin{table*}[t]
    \centering
    \begin{tabular}{c}
    \includegraphics[width=0.98\textwidth]{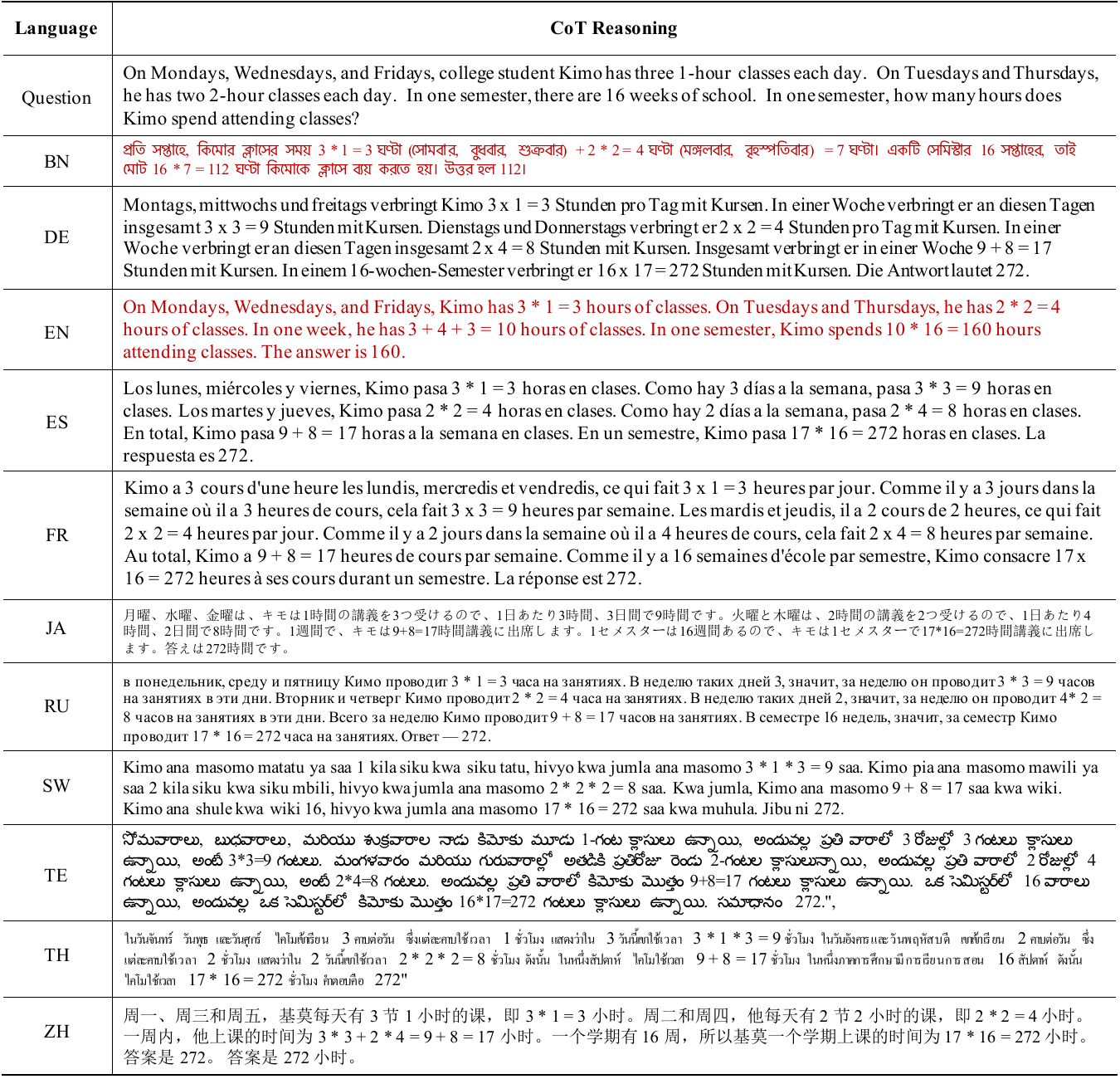}
    \end{tabular}
    \caption{Case study from the MGSM test set, where solutions are generated by Qwen2.5-72B for the same question written in different languages. Note that, for brevity, we only show the English (EN) question here; the corresponding questions in the other languages are provided in Appendix~\ref{app:question}.}
    \label{tab:example}
\end{table*}

\subsection{Analysis}

\paragraph{Accuracy–efficiency Trade-off}
One can conceive CLC as a multilingual extension of monolingual self-consistency, going beyond the typical sampling-based strategy. 
Figure~\ref{fig:varying-paths} reports accuracy under different numbers of reasoning paths for Llama3.1 and Qwen2.5. For self-consistency, we use English following~\citep{wang2023selfconsistency}, whereas for CLC, we use 8 languages excluding BN, SW, and TE which have very low results (see Table~\ref{tab:clc-gsm8k}). We sample $N/8$ reasoning paths for each language, thus creating $N$ solutions for CLC. 
We see that Llama3.1 shows CLC slightly below self-consistency but above greedy decoding, while Qwen2.5 achieves notable gains with CLC, surpassing self-consistency when $N\geq16$. 
These results suggest that cross-lingual diversity provides complementary reasoning signals. However, as with self-consistency, this improvement comes at the expense of higher computational effort, as our methods require generating multiple solutions across languages for CLC and analogously for COC and CPC. For instance, models generally require generating 8–16 reasoning solutions to surpass greedy decoding, which corresponds to roughly 8–16$\times$ the number of generated reasoning paths, and thus substantially higher inference cost. This highlights the inherent trade-off between accuracy and efficiency: while additional paths and languages lead to better reasoning, they also increase inference overhead.


\paragraph{Case Study}
Table~\ref{tab:example} shows examples of solutions generated by Qwen2.5-72B for the same MGSM question  in different languages. For the EN question, the model incorrectly reasons in the first step: \textit{On Mondays, Wednesdays, and Fridays, Kimo has 3 $*$ 1 = 3 hours of classes}, which results in the wrong reasoning and final answer. It is interesting to see that the model reasons correctly in the other languages except BN, while the solution paths might be logically different. For DE, the model first calculates the total hours for Monday, Wednesday, and Friday, then calculates the total hours for Tuesday and Thursday, and then adds them together to get the total hours for the week; while for ZH, it calculates the hours per day and then calculates the total hours for the week. Reasoning appears correct in some low-resource languages (SW, TE, and TH), suggesting that our approach can leverage the model’s capabilities in low-resource languages to assist reasoning in high-resource languages, in addition of course to the usual reverse benefit.

\paragraph{Correlation}
To further understand the relationship between reasoning performance and reasoning consistency, we analyse their correlations under different variations (Figure~\ref{fig:cor}). We can observe that COC and CPC have high correlations with reasoning accuracy, while  CLC has a weak and non-significant one. This suggests that we can use COC and CPC to assess the model's uncertainty in its generated solutions without using gold answers. While CLC does not appear to be a reliable proxy for model accuracy in our experiments, it can still be used to evaluate models from a multilingual perspective.

\section{Conclusion}
MRC is a multidimensional reasoning framework, which comprehensively studies reasoning consistency when the same math problems are presented to the model with systematic variations along three different dimensions. 
By leveraging such variations and answer consistency, MRC improves overall accuracy on both monolingual and multilingual benchmarks, demonstrating its effectiveness in enhancing lightweight models and its strong scalability for more complex tasks.
Our experiments seem to suggest that the greater the diversity of solution paths, the stronger the benefit from exploiting consistency. As we do not yet have concrete evidence for this hypothesis, a natural future direction would be to study path diversity in a quantifiable way. 
Another valid extension would be \textit{integrating} the different  dimensions (e.g., COC per language).  While combining multiple dimensions of variations 
presents an explosion of possibilities, strategic selection based on empirical results, especially cross-all accuracy, might mitigate this challenge.

\section{Limitations}
While we investigated model consistency in mathematical reasoning and successfully leveraged it to improve reasoning accuracy, several promising directions remain for future exploration. We mainly focus on the variations in model inputs and consistency in final answers, while both the variation and consistency of the reasoning paths are interesting directions. Specifically, variations in the input will lead to variations in the output, which includes logical consistency and inconsistency, thus affecting the final result. Lastly, it is not yet clear how variations in input affect the model's reasoning logic (variations in reasoning paths), which requires a much bigger unpacking. 


\section*{Acknowledgments}
The anonymous reviewers of ACL Rolling Review provided us with useful comments which contributed to improving this paper and its presentation, so we’re grateful to them. We also gratefully acknowledge funding from the Dutch Ministry through the Sector Plan in the Humanities ``Humane AI.'' This work used the Dutch national e-infrastructure with the support of the SURF Cooperative using grant no. EINF-14851.


\bibliography{custom}

\clearpage
\appendix
\onecolumn
\section{Appendix}
\label{sec:appendix}

\subsection{Prompt Examples}
\label{sub:prompt}

\begin{figure*}[ht]
    \centering
    \includegraphics[scale=.52]{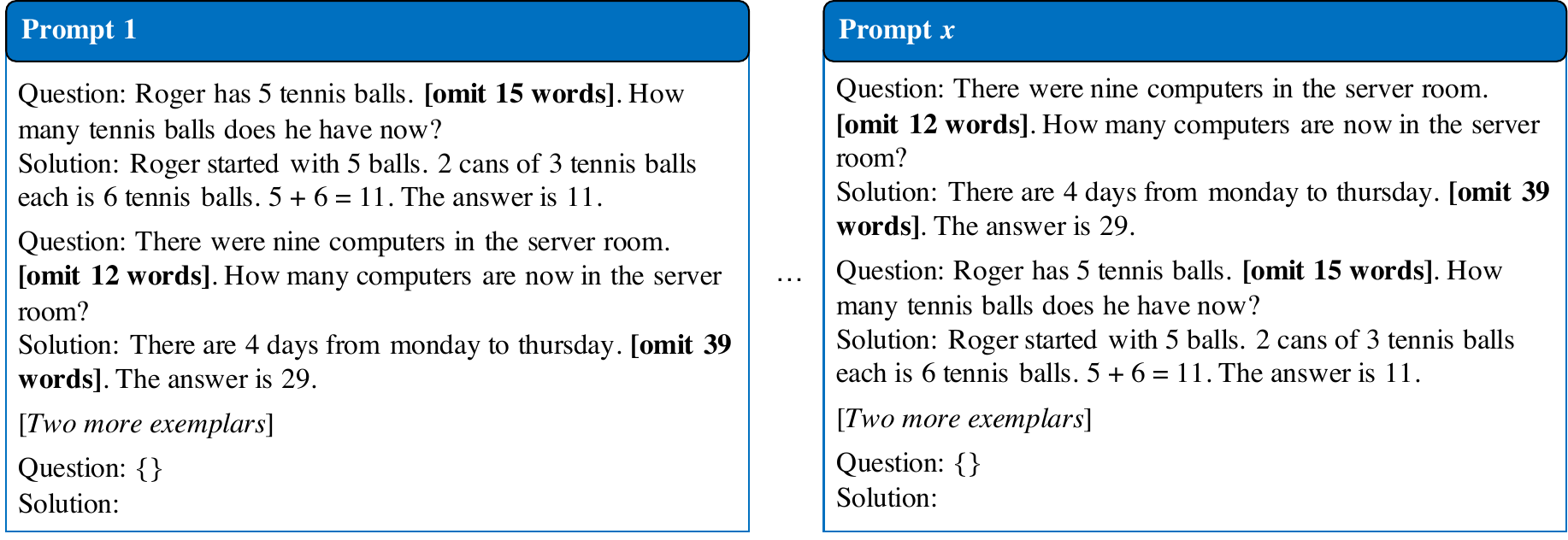}
\caption{Examples of prompts for COC.} 
\label{fig:coc-prompt}
\end{figure*}

\begin{figure*}[ht]
    \centering
    \includegraphics[scale=.52]{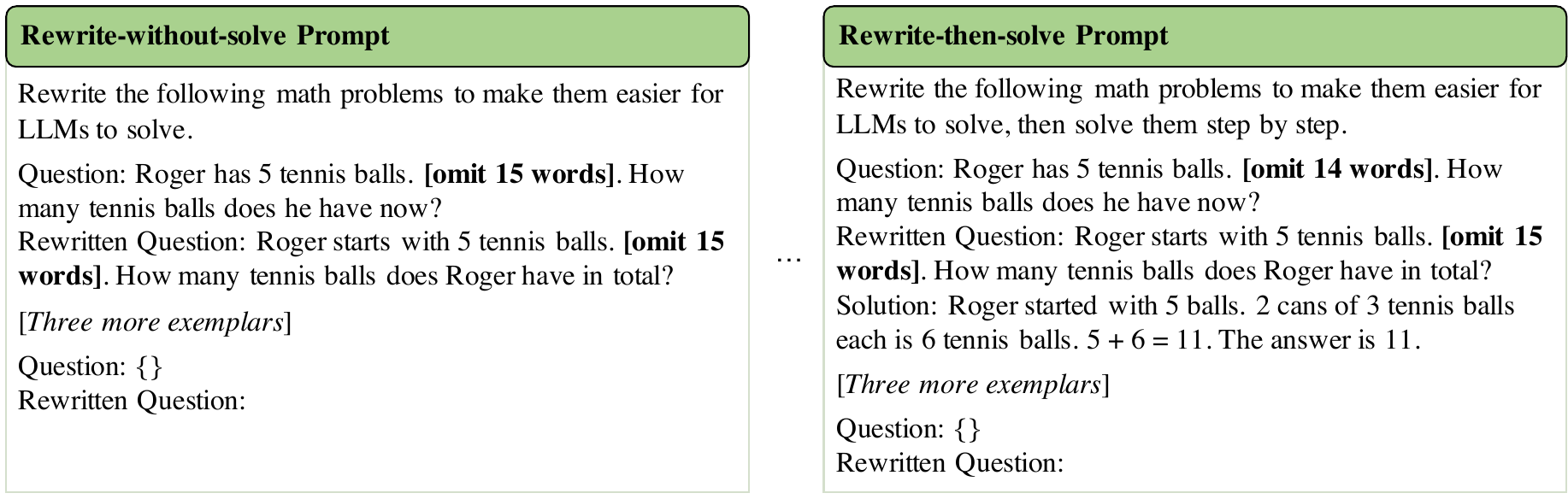}
\caption{Examples of prompts for CPC.} 
\label{fig:cpc-prompt}
\end{figure*}

\begin{figure*}[!ht]
    \centering
    \includegraphics[scale=.52]{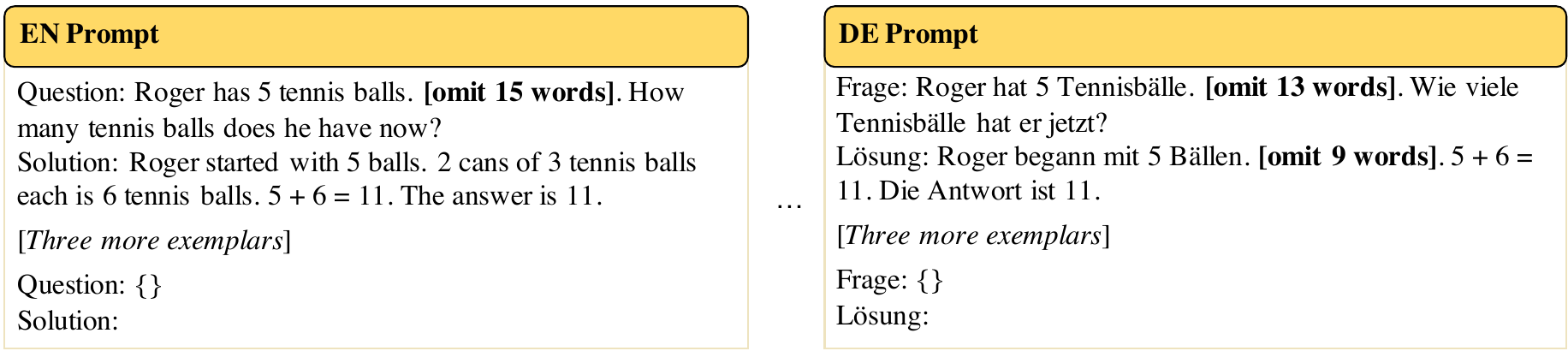}
\caption{Examples of prompts for CLC.} 
\label{fig:clc-prompt}
\end{figure*}

\newpage

\subsection{Models}
\label{app:models}

We select a range of open-source state-of-the-art LLMs in varying scales. For all models, we only consider instruction-tuned versions. 

\paragraph{7-8B:} Phi3-7B (128k)~\citep{marah-etal-2024-phi3}; Qwen2.5-7B~\citep{yang-etal-2024-qwen2}; Qwen2.5-Math-7B~\cite{yang-etal-2024-qwen2.5math}; Llama3.1-8B~\citep{dubey-etal-2024-llama3}; Aya-expanse-8B~\citep{ahmet-etal-2024-aya}; Ministral-8B\footnote{\url{https://huggingface.co/mistralai/Ministral-8B-Instruct-2410}}.

\paragraph{14-32B:} Phi3-14B~\citep{marah-etal-2024-phi3}; Mistral-22B\footnote{\url{https://huggingface.co/mistralai/Mistral-Small-Instruct-2409}}; Aya-expanse-32B~\citep{ahmet-etal-2024-aya}.

\paragraph{70-72B:} Qwen2.5-72B~\citep{yang-etal-2024-qwen2}; Qwen2.5-Math-72B~\cite{yang-etal-2024-qwen2.5math}; Llama3.1-70B~\citep{dubey-etal-2024-llama3}.


\subsection{COC Results}
\label{tab:coc-results}

\begin{table*}[ht]
\centering
\footnotesize
\setlength{\tabcolsep}{6pt}
\begin{tabular}{lrrrrrrrrr}
\toprule
\textbf{Orders (4-shot)} &  \makecell[c]{\textbf{1}} & \makecell[c]{\textbf{2}} & \makecell[c]{\textbf{3}} & \makecell[c]{\textbf{4}} & \makecell[c]{\textbf{5}} & \makecell[c]{\textbf{6}} & \makecell[c]{\textbf{7}} & \makecell[c]{\textbf{8}} & \makecell[c]{\textbf{COC}}\\
\hline
7-8B\\

Phi3-7B    & 88.5 & 88.8 & 88.6 & 88.9 & 87.3 & 89.8 & 88.4 & 88.2 & 89.8\\
Qwen2.5-7B  & 88.3 & 87.0 & 87.6 & 88.0 & 88.0 & 87.8 & 88.5 & 88.2 & 90.5\\
Qwen2.5-Math-7B & 90.0 & 89.7 & 91.3 & 92.3 & 91.5 & 90.5 & 90.8 & 91.0 & 92.6\\
Llama3.1-8B & 79.7 & 78.8 & 79.0 & 79.0 & 80.1 & 79.1 & 77.8 & 78.7 & 80.1\\
Aya-expanse-8B & 76.7 & 77.3 & 78.2 & 76.6 & 76.9 & 77.6 & 77.3 & 77.4 & 78.2\\
Ministral-8B & 81.2 & 81.4 & 80.9 & 81.4 & 80.9 & 81.3 & 81.4 & 81.6 & 82.3\\
\hline
14-32B\\
Phi3-14B   & 89.2 & 88.9 & 88.6 & 88.3 & 88.9 & 89.0 & 89.1 & 88.9 & 89.9\\
Mistral-22B & 85.8 & 85.8 & 85.9 & 85.3 & 86.1 & 86.1 & 86.1 & 86.2 & 86.7\\
Aya-expanse-32B & 83.8 & 83.4 & 84.6 & 83.6 & 84.9 & 84.3 & 83.5 & 83.9 & 85.3\\
\hline
70-72B\\
Llama3.1-70B & 94.0 & 94.0 & 94.3 & 94.1 & 93.9 & 93.9 & 93.4 & 93.8 & 94.4
\\
Qwen2.5-72B & 94.6 & 93.9 & 93.9 & 93.6 & 94.1 & 94.1 & 94.2 & 94.2 & 94.8\\
Qwen2.5-Math-72B & 94.0 & 93.6 & 93.6 & 93.7 & 93.8 & 93.6 & 93.2 & 93.4 & 94.2\\
\bottomrule
\end{tabular}
\caption{\label{tab:coc-gsm8k}
Reasoning accuracy of prompts in different orders on GSM8K compared to COC.
}
\end{table*}

\begin{table*}[ht]
\centering
\footnotesize
\setlength{\tabcolsep}{6pt}
\begin{tabular}{lrrrrrrrrr}
\toprule
\textbf{Orders (4-shot)} &  \makecell[c]{\textbf{1}} & \makecell[c]{\textbf{2}} & \makecell[c]{\textbf{3}} & \makecell[c]{\textbf{4}} & \makecell[c]{\textbf{5}} & \makecell[c]{\textbf{6}} & \makecell[c]{\textbf{7}} & \makecell[c]{\textbf{8}} & \makecell[c]{\textbf{COC}}\\
\hline
7-8B\\

Phi3-7B        & 40.4 & 41.0 & 40.0 & 40.8 & 39.0 & 42.2 & 40.0 & 38.6 & 43.2\\
Qwen2.5-7B      & 63.6 & 64.6 & 64.8 & 64.4 & 63.6 & 65.2 & 61.8 & 62.4 & 68.8\\
Qwen2.5-Math-7B & 75.2 & 74.4 & 75.2 & 75.6 & 73.2 & 75.4 & 74.6 & 74.2 & 77.0\\
Llama3.1-8B    & 42.6 & 44.0 & 41.8 & 41.4 & 42.4 & 41.6 & 45.0 & 42.6 & 48.0\\
Aya-expanse-8B  & 28.0 & 28.8 & 29.0 & 27.4 & 28.6 & 28.8 & 27.0 & 29.0 & 31.4\\
Ministral-8B    & 42.8 & 42.2 & 40.0 & 41.4 & 42.0 & 41.2 & 41.8 & 41.2 & 45.0\\
\hline
14-32B\\
Phi3-14B       & 44.2 & 43.0 & 44.2 & 45.8 & 44.4 & 46.6 & 44.0 & 44.4 & 49.0\\
Mistral-22B     & 42.6 & 47.2 & 43.0 & 44.0 & 44.0 & 44.6 & 41.6 & 45.6 & 48.2\\
Aya-expanse-32B & 38.6 & 40.0 & 39.4 & 38.4 & 40.4 & 40.0 & 40.4 & 40.6 & 43.4\\
\hline
70-72B\\
Llama3.1-70B   & 58.6 & 57.8 & 60.0 & 59.6 & 59.0 & 60.2 & 60.2 & 61.4 & 65.0\\
Qwen2.5-72B     & 63.2 & 65.2 & 62.2 & 59.6 & 67.2 & 61.4 & 69.8 & 68.6 & 75.2\\
Qwen2.5-Math-72B& 66.6 & 64.6 & 70.2 & 72.2 & 68.8 & 69.6 & 72.6 & 69.8 & 76.0\\
\bottomrule
\end{tabular}
\caption{\label{tab:coc-math500}
Reasoning accuracy of prompts in different orders on MATH500 compared to COC.
}
\end{table*}

\subsection{Examples of questions written in different languages}

\begin{figure*}[ht]
    \centering
    \includegraphics[scale=.7]{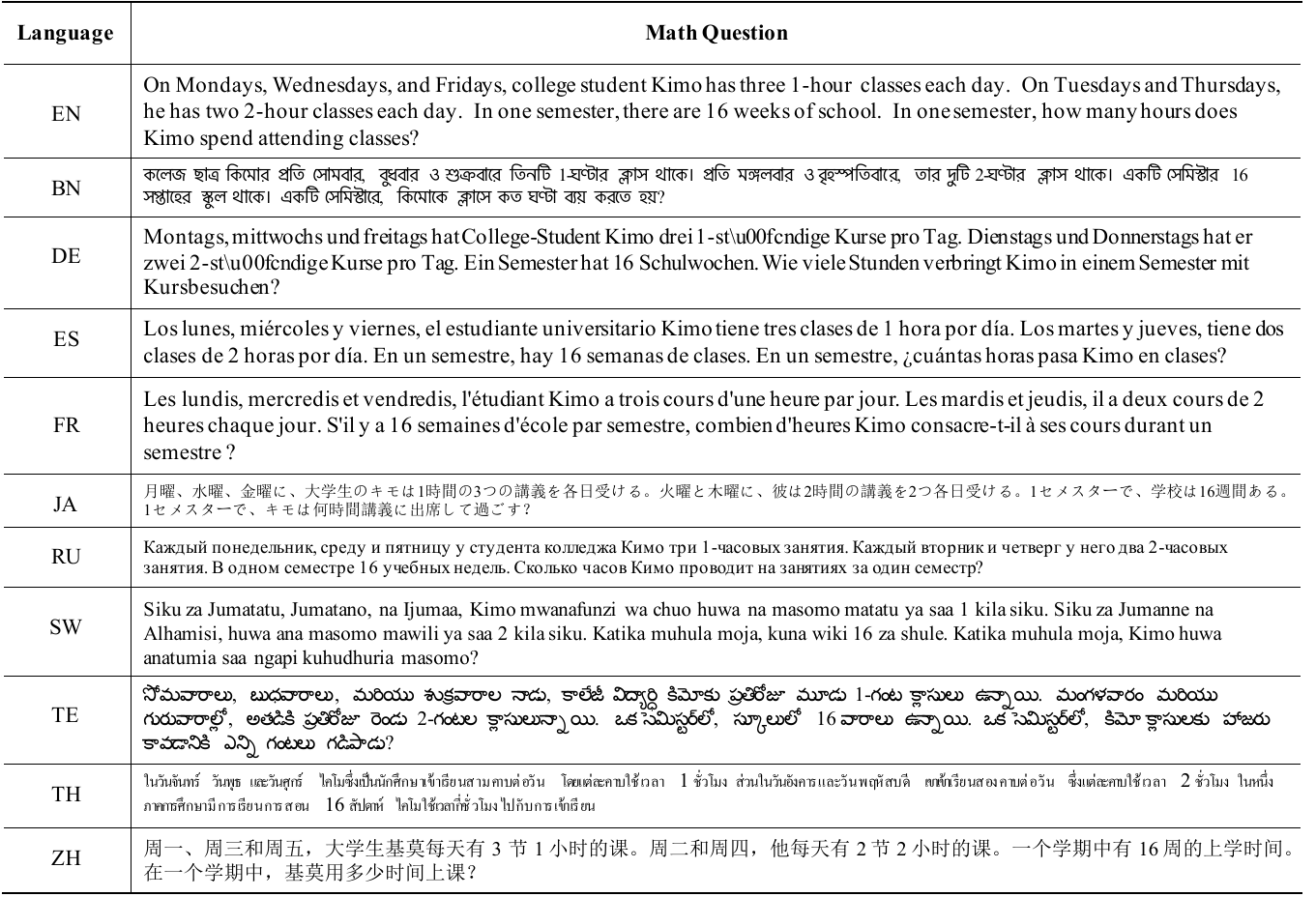}
\caption{Examples of questions written in different languages.} 
\label{fig:questions}
\end{figure*}

Figure~\ref{fig:questions} provides examples of questions written in different languages.

\label{app:question}





\end{document}